\documentclass[runningheads,a4pasper]{llncs}

\usepackage{amssymb}
\usepackage{graphicx}

\usepackage{url}
\urldef{\mailsa}\path|{sbeliga, amestrovic, smarti}@uniri.hr|   
\newcommand{\keywords}[1]{\par\addvspace\baselineskip
\noindent\keywordname\enspace\ignorespaces#1}

\begin{document}

\mainmatter

\title{Toward Selectivity Based Keyword Extraction for Croatian News}

\titlerunning{Toward Selectivity Based Keyword Extraction for Croatian News}

\author{Slobodan Beliga, Ana Me\v{s}trovi\'c, Sanda Martin\v{c}i\'c-Ip\v{s}i\'c}

\authorrunning{Slobodan Beliga, Ana Me\v{s}trovi\'c, Sanda Martin\v{c}i\'c-Ip\v{s}i\'c}

\institute{Department of Informatics,\\
University of Rijeka,\\
Radmile Matej\v{c}i\'c 2, 51000 Rijeka, Croatia \\
\mailsa\\
}

\maketitle

\begin{abstract}
Preliminary report on network based keyword extraction for Croatian is an unsupervised method for keyword extraction from the complex network. We build our approach with a new network measure - the node selectivity, motivated by the research of the graph based centrality approaches. The node selectivity is defined as the average weight distribution on the links of the single node. We extract nodes (keyword candidates) based on the selectivity value. Furthermore, we expand extracted nodes to word-tuples ranked with the highest in/out selectivity values. Selectivity based extraction does not require linguistic knowledge while it is purely derived from statistical and structural information encompassed in the source text which is reflected into the structure of the network. Obtained sets are evaluated on a manually annotated keywords: for the set of extracted keyword candidates average $F1$ score is 24,63\%, and average $F2$ score is 21,19\%; for the exacted words-tuples candidates average $F1$ score is 25,9\% and average $F2$ score is 24,47\%.

\keywords{keywords extraction, complex networks, Croatian texts, selectivity}
\end{abstract}

\section{Introduction}

The task of keyword extraction is to automatically identify a set of terms that best describe the document \cite{mihalcea2004textrank}. Automatic keyword extraction establishes foundation for various natural language processing applications: information retrieval, automatic indexing and classification of documents, automatic summarization, high-level semantic description, etc. 

State-of-the-art keyword extraction approaches are based on statistical methods which require learning from hand-annotated data sets. In the last decade focus of research has shifted toward unsupervised methods, mainly toward network or graph enabled keyword extraction. In a network based keyword extraction the document representation may vary from very simple (words are nodes and their co-occurrence is represented with links), or can incorporate very sophisticated linguistic knowledge like syntactic \cite{liu2008role} or semantic relations \cite{tsatsaronis2010semanticrank}. Typically, the source (document, text, data) for keyword extraction is modelled with one network. This way, both the statistical properties (frequencies) as well as the structure of source text are represented by unique formal representation, hence complex network.    

Network (or graph, since the number of words in isolated documents is limited) enabled keyword extraction exploits different measures for the task of identification and ranking of the most representative features of the source -  the keywords. The keyword extraction powered by  network measures can be on the node, network or subnetwork level. Measures on the node level are: degree, strength, centrality \cite{lahiri2014keyword}; on the network level: coreness, clustering coefficient, PageRank motivated ranking score or HITS motivated hub and authority score \cite{boudin2013comparison}\cite{litvak2008graph}\cite{mihalcea2004textrank}; on the subnetwork level: communities \cite{grineva2009extracting}. Majority of the research was motivated with various centrality measures: degree, betweenness, closeness and eigenvector centrality \cite{lahiri2014keyword}\cite{boudin2013comparison}\cite{litvak2008graph}\cite{palshikar2007keyword}\cite{mihalcea2004textrank}\cite{matsuo2001keyworld}.

Our research aims at proposing an unsupervised method for keyword extraction in Croatian language. Since, Croatian is a highly flective Slavic language, usually source text needs a substantial preprocessing (lemmatization – morphological normalization, stopwords removal, part-of-speech (POS) annotation, morphosyntactic descriptions (MSD)tagging, etc.), we design our approach with light or no linguistic knowledge. A new network measure - the node selectivity, originally proposed by Masucci and Rodgers \cite{masucci2006network} \cite{masucci2009differences} (which that can distinguish a real  from a shuffled network), is applied to automatic keyword extraction. 
Selectivity is defined as the average weight distribution on the links of the single node. In our previous work, the node selectivity measure performed in favour of differentiation between original and shuffled Croatian texts \cite{margan2014shuffled}\cite{margan2014shuffled2}, and for differentiation of blog and literature text genres \cite{sisovic2014blogs}. To the best of our knowledge, the node selectivity measure was not applied to keyword extraction task before.

Section 2 presents an overview of related work on automatic keyword extraction. In Section 3 we present definition of the measures for the network structure analysis. In Section 4 we present the construction of co-occurrence networks from used text collection. In Section 5 are the results of keyword extraction, and in the final Section, we elaborate the obtained results and make conclusions regarding future work.

\section{Related work}
Lahiri et al. \cite{lahiri2014keyword} extract keywords and keyphrases form co-occurrence networks of words and from noun phrases collocations networks. Eleven measures (degree, strength, neighbourhood size, coreness, clustering coefficient, structural diversity index, page rank, HITS – hub and authority score, betweenness, closeness and eigenvector centrality) are used for keyword extraction from directed/undirected and weighted networks. Obtained results on 4 data sets are suggesting that centrality measures outperform the baseline term frequency/inverse document frequency (tf-idf) model, and simpler measures like degree and strength outperform computationally more expensive centrality measures like coreness and betweenness. 

Boudin \cite{boudin2013comparison} compare various centrality measures for graph-based keyphrase extraction. Experiments on standard datasets of English and French show that simple degree centrality achieve results comparable to the widely used TextRank algorithm;  and that closeness centrality obtains the best results on short documents. Undirected and weighted co-occurrence networks are constructed from syntactically (only nouns and adjectives) parsed and lemmatized text using co-occurrence window. Degree, closeness, betweenness and eigenvector centrality are compared to PageRank proposed by Mihalcea in \cite{mihalcea2004textrank} as a baseline. Degree centrality achieve similar performance as much complex TextRank and closeness centrality outperforms TextRank on short documents (scientific papers abstracts).

Litvak and Last \cite{litvak2008graph} compare supervised and unsupervised approach for keywords identification in the task of extractive summarization. Approaches are based on the graph-based syntactic representation of text and web documents. The results of the HITS algorithm on a set of summarized documents performed comparable to supervised methods (Naive Bayes, J48, Support Vector Machines). Authors are suggesting that simple degree-based rankings from the first iteration of HITS, rather than running it to its convergence should be considered. 

Grineva et al. \cite{grineva2009extracting} use community detection techniques for key terms extraction on Wikipedia's texts, modelled as a graph of semantic relationships between terms. The results showed that the terms related to the main topics of the document tend to form a community, thematically cohesive groups of terms. Community detection allows effectively processing of multiple topics in document and efficiently filters out noise. The results achieved on weighted and directed networks from semantically linked, morphologically expanded and disambiguated n-grams from article's titles. Additionally, for the purpose of the noise stability, they repeated the experiment on different multi-topic web pages (news, blogs, forums, social networks, product reviews) which confirmed the community detection outperforms td-idf model. 

Palshikar \cite{palshikar2007keyword} propose a hybrid structural and statistical approach to extract keywords from a single document. The undirected co-occurrence network, using a dissimilarity measure between two words, calculated from the frequency of their co-occurrence in the preprocessed and lemmatized document, as the edge weight was shown to be appropriate for centrality measures based approach for keyword extraction. 

Mihalcea and Tarau \cite{mihalcea2004textrank} report a seminal research which introduced state-of-the-art TextRank model. TextRank is derived from PageRank and introduced to  graph based text processing, keyword and sentence extraction. Abstracts are modelled as undirected or directed and weighted co-occurrence networks using the co-occurrence window of variable sizes (2..10). Lexical units are preprocessed: stopwords removed, words restricted with POS syntactic filters (open class words, nouns and adjectives, nouns). The PageRank motivated score of importance of the node derived from importance of the neighboring nodes is used for keyword extraction. Obtained TextRank performance compares favorably with the supervised machine learning n-gram based approach. 

 Matsou et al. in \cite{matsuo2001keyworld} present an early research which represents a document by a undirected and unweighted co-occurrence network, and study small-world properties of text. Based on the network topology, the authors proposed an indexing system called KeyWorld, which extracts important terms (pair of words) by measuring their contribution to small world properties. The contribution of the node is based on closeness centrality calculated as the difference in small world properties of the network with temporarily elimination of a node combined with inverse document frequency (idf).  

Erkan and Radev \cite{erkan2004lexrank} introduce a stochastic graph-based method for computing relative importance of textual units on the problem of text summarization by extracting the most important sentences. LexRank, calculates sentence importance based on the concept of eigenvector centrality in a graph representation of sentences. A connectivity matrix based on intra-sentence cosine similarity is used as the adjacency matrix of the graph representation of sentences. LexRank is shown to be quite insensitive to the noise in the data. 

Mihalcea in \cite{mihalcea2004graph} presents extension to earlier work \cite{mihalcea2004textrank}, where the TextRank algorithm is applied for text summarization task powered by sentence extraction. On this task TextRank performed at par with the supervised and unsupervised summarization methods, which motivated the new branch of research based on the graph-based extracting and ranking algorithms.

Tsatsaronis et al. \cite{tsatsaronis2010semanticrank} present SemanticRank, a network based ranking algorithm for keyword and sentence extraction from text. Semantic relation is based on calculated knowledge-based measure of semantic relatedness between linguistic units (keywords or sentences). The keyword extraction from the Inspec abstracts results reported favorable performance of SemanticRank over state-of-the-art counteparts - weighted and unweighted variations of PageRank and HITS. 

Huang et al. \cite{huang2006keyphrase} propose an automatic keyphrase extraction algorithm using unsupervised method based on conectedness and beetweenness centrality. 

\subsection{Related work on Croatian langauge}
The keyphrase extraction for Croatian language has been addressed in both supervised \cite{ahel2009automatic} and unsupervised \cite{mijic2010robust}\cite{bekavac2013gpkex}\cite{saratlija2011unsupervised} settings. Ahel et al. \cite{ahel2009automatic} use a Naive Bayes classifier combined with tf-idf (term frequency/inverse document frequen-cy), \cite{mijic2010robust} use part-of-speech (POS) and morphosyntactic description (MSD) tags filtering followed by tf-idf ranking, and \cite{saratlija2011unsupervised} use distributional semantics to build topically related word clusters, from which they extract keywords and expand them to keyphrases. Bekavac et al. \cite{bekavac2013gpkex} propose a genetic programming approach for keyphrases extraction for Croatian language on the same data set. GPKEX can evolve simple and interpretable keyphrase scoring measures that perform comparably to other machine learning methods for Croatian. Reported research on extraction of Croatian keywords use data set composed of Croatian news articles from the Croatian News Agency (HINA), with hand annotated keywords by human experts.

\section{The complex network analysis}

In the network, $N$ is the number of nodes and  $K$ is the number of links. In weighted language networks every link connecting two nodes $i$ and $j$ has an associated weight $w_{ij}$ that is a positive integer number \cite{newman2010networksbook}.

For every two connected nodes $i$ and $j$ the number of links lying on the shortest path between them is denoted as $d_{ij}$. For every two connected nodes $i$ and $j$ the number of shortest paths from node  $j$ to node $k$ is denoted as $\sigma_{jk}$, while the number of those paths that pass through node $i$ is denoted as  $\sigma_{jk}(i)$.

The node degree $k_{i}$ is defined as the number of edges incident upon a node. The in-degree and out-degree $k_{i}^{in/out}$ of node $i$ is defined as the number of its in and out neighbours.

Degree centrality of the node $i$ is the degree of that node normalised by dividing by the maximum possible degree N-1 \cite{newman2010networksbook}:

\begin{equation}
dc_{i} =  \frac{k_{i}}{N-1}.
\end{equation} 

Analogue, the in/out-degree centralities are defined as in/out – degree of a node:

\begin{equation}
dc_{i}^{(in/out)} =  \frac{k_{i}^{(in/out)}}{N-1}.
\end{equation} 

Closeness centrality is defined as the inverse of farness, i.e. the sum of the shortest distances between a node and all the other nodes. The normalised closeness centrality of a node $i$ is given by \cite{newman2010networksbook}:

\begin{equation}
cc_{i} =  \frac{N-1}{\sum_{i \neq j}d_{ij}}.
\end{equation} 

Betweenness centrality quantifies the number of times a node acts as a bridge along the shortest path between two other nodes. In the directed network, the normalised betweenness centrality of a node $i$ is given by \cite{newman2010networksbook}:

\begin{equation}
bc_{i} =  \frac{\sum_{i \neq j \neq k}\frac{\sigma_{jk}(i)}{\sigma_{jk}}}{(N-1)(N-2)}.
\end{equation}  

The strength of the node $i$ is a sum of weights of all links incident with the node $i$:

\begin{equation}
s_{i} =  \sum_{j}w_{j}.
\end{equation}

In the directed network, the in/out-strength $s_{i}^{in/out}$ of the node $i$ is defined as the number of its incoming and outgoing  links, that is: 

\begin{equation}
s_{i}^{in/out} =  \sum_{j}w_{ji/ij}.
\end{equation} 

The selectivity of the node $i$ is a fraction of the node weight and node degree \cite{masucci2009differences}:

\begin{equation}
e_{i} = \frac{s_{i}}{k_{i}}.
\end{equation}

In the directed network, the in/out selectivity of the node $i$ is defined as:

\begin{equation}
e_{i}^{in/out} = \frac{s_{i}^{in/out}}{k_{i}^{in/out}}.
\end{equation}

\section{Methodology}
\subsection{Data}

For the network based keyword extraction we use the data set composed of Croatian news articles \cite{mijic2010robust}. The dataset contains 1020 news articles from the Croatian News Agency (HINA), with hand annotated keywords by human experts. The set is divided: 960 annotated documents for learning of supervised methods, and 60 documents for testing. The test set of 60 documents is annotated by 8 different experts, where the inter-annotator agreement in terms of F2 scores (see section 6) are in average 46\% (between 29.3\% and 66.1\%). 

We selected the first 30 texts from the HINA collection for our preliminary experiment. The texts required some preprocessing: parsing only textual part and title part excluding annotations, cleaning of diacritics and symbols (w instead vv, ! instead l, etc.) and lemmatization. Non-standard word forms  numbers, dates, acronyms, abbreviations etc. remain in text, since the method is preferably resistant to the noise presented in the data source. The selected 30 texts varied in the length: from very short 60 tokens up to 800 tokens (318 tokens in average).

\subsection{The construction of co-occurrence networks}

Text can be represented as a complex network of linked words: each individual word is a node and interactions amongst words are links. Co-occurrence networks exploit simple neighbour relation, two words are linked if they are adjacent in the sentence \cite{margan2013preliminary}. The weight of the link is proportional to the overall co-occurrence frequencies of the corresponding word pairs within a corpus. For each document in HINA dataset we construct one directed and weighted co-occurence network, 30 in total. 

Network construction and analysis was implemented with the Python programming language using the NetworkX software package developed for the creation, manipulation, and study of the structure, dynamics, and functions of complex networks \cite{hagberg2008exploring}. 

\subsection{Selectivity-based keyword extraction}

Selectivity is a local (node level) network measures. In weighted and directed co-occurrence networks one can consider the in- and out- links for obtaining in/out selectivity of the node (Eq.8).

In the first part of our experiment we compute in/out selectivity for all nodes in all 30 networks. The nodes are ranked according to the highest in/out selectivity values above threshold value. Preserving the same threshold value ($geq$ 1) in all documents resulted in different number of nodes (one word long keyword candidates) extracted per each network. Obtained set of one word long keyword candidates is noted as SET1.  

Then, for every filtered node we detect neighbouring nodes: for the in-selecti-vity we isolate one neighbour node with the highest outgoing weight; for the out-selectivity we isolate one neighbour node with the highest ingoing weight. The result of in/out selectivity extraction is a set of ranked words-tuples - SET2. Words-tuples are  two words long sequences of keyword candidates. From obtained tuples we filtered out those containing stopwords in order to compare with the manually annotated evaluation set.

\section{Evaluation and Results}

For the keyword extraction task the strategy "more is better" can be utilized, since there is no objective judgement on keywords. Hence, it is preferable to extract more keywords which makes trade off between precision and recall of the methods. The second polemic issue of keyword extraction task is: shorter keywords are more general vs. longer which are more accurate. Motivated by this open arguments, and by the approach of other authors, we decided to follow the same principle: to extract as many as possible keyword candidates and evaluate them on the basis of recall ($R$) and $F2$ score, beside the standard precision ($P$) and $F1$ score. $F1$ score is a harmonic mean of precision and recall: $F_{1}=2PR/(P+R)$. $F2$ measure gives two times as much importance to recall as to precision: $F_{2}=5PR/(4P+R)$. 

Evaluation is the final part of the experiment based on the intersection of  the obtained sets SET1 and SET2 of keyword candidates with the union of all 8 annotators keywords. The results in terms of precision and recall are in Figures 1 and 2 respectively, and in terms of $F1$ and $F2$ scores in Figures 3 and 4 respectively. Obtained average $F1$ score for the SET 1 is 24,63\%, and average $F2$ score is 21,19\%. Expansion of obtained candidates to SET2 increased average $F1$ score to 25,9\% and $F2$ score to 24,47\%.

\begin{figure}
\centering 
\includegraphics[width=\textwidth]{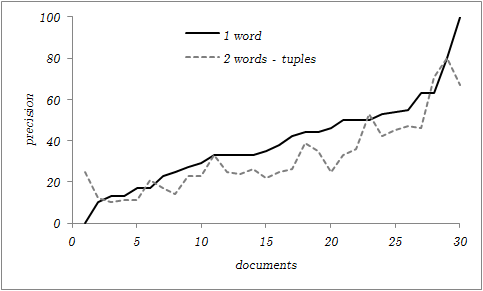}
\caption{Precision on the SET1 (1 word candidates) and SET2 (2 words-tuples candidates) per 30 documents}
\end{figure}

\begin{figure}
\centering 
\includegraphics[width=\textwidth]{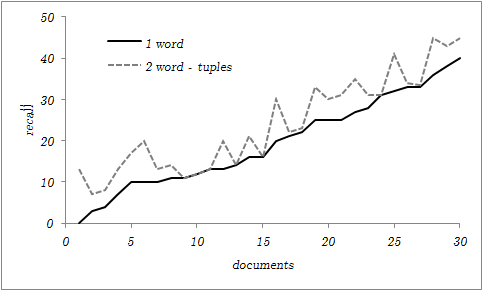}
\caption{Recall on the SET1 (1 word candidates) and SET2 (2 words-tuples candidates) per 30 documents}
\end{figure}

All supervised and unsupervised methods reported on keyphrases extraction from HINA dataset incorporate linguistic knowledge (POS, MSD,..) of Croatian.  Miji\'c  at al. \cite{mijic2010robust} initially extracted the list of keyword candidates as a comprehensive list of all words without stopwords) which was expanded in a longer n-gram sequences up to length 4. In \cite{mijic2010robust} a keyphrase extraction system developed for a large-scale Croatian news production system the tf-idf ranking model was used to extract n-grams of up to length four, which were lemmatized, and POS and MSD filtered. For evaluation the manually annotated key phrases from 60 documents were used. The evaluation set was reduced to keywords suggested only by 3 top annotators (have the highest inter-annotator agreement among all 8 annotators). The results indicate that the performance is comparable to that of the human annotators. Ahel at al. supervised \cite{ahel2009automatic} for the one word long keywords reported precision of 22\% and recall of 3.4\%. 

We designed our method purely from statistical and structural information encompassed in the source text which is reflected into the structure of the network. Our method achieved on a SET1 average recall of 19,53\% and precision of 39,1\%. Expansion to the words-tuples in SET2 increased average recall to 23,87\% and decreased precision to 32,23\%. Obtained results are comparable to \cite{mijic2010robust} and \cite{ahel2009automatic}, but with slightly different evaluation set up. 

The preliminary results are promising and have potential to improve in several directions which is elaborated at the end of the next section. Additional remark regarding results, is that beside keyword candidates our method captures personal names and entities, which were not marked as keyphrases and lowered the score. Capturing names and entities can be of high relevance for the tasks like name-entity recognition, text summarization, etc. 

\begin{figure}
\centering 
\includegraphics[width=\textwidth]{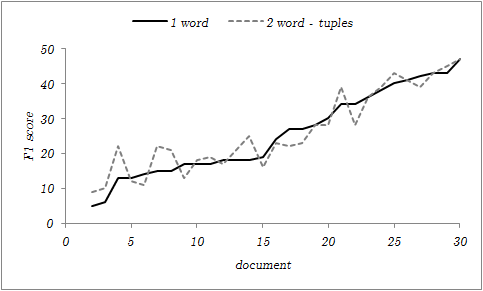}
\caption{F1 score of the SET1 (1 word candidates) and SET2 (2 words-tuples candidates) per 30 documents }
\end{figure}

\begin{figure}
\centering 
\includegraphics[width=\textwidth]{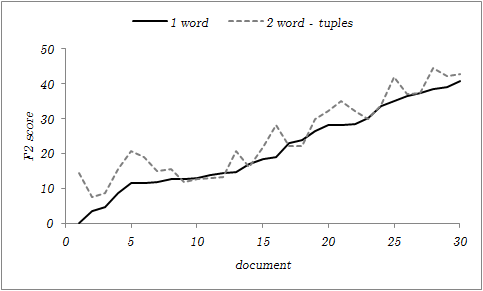}
\caption{F2 score of the SET1 (1 word candidates) and SET2 (2 words-tuples candidates) per 30 documents }
\end{figure}

Keyword annotation is an extremely subjective task as even human experts have difficulties to agree upon keyphrases (inter-agreement around 40\%). Croatian is highly morphologically rich language, which puts another magnitude of challenge on the task, since annotators are freely choosing morphological word form as tag, which seems appropriate at the moment. Additionally, there was no predefined set of index or keywords list, so annotators could make up their own, even worse in some cases it seemed appropriate to annotate with keywords, which were not present in the original article (out-of-vocabulray words). In \cite{ahel2009automatic} the number of out-of-vocabulary keywords on the whole HINA dataset is estimated to high 57\%. Since our method is derived from purely text statistics, it is not capable to capture all possible subjective variations of annotators.

\section{Conclusion}

Preliminary report on selectivity based keyword extraction for Croatian news describes an unsupervised method which extracts nodes from complex network as keyword candidates. We build our approach with a new network measure - the node selectivity (defined as the average weight distribution on the links of the single node), motivated by the research of the graph-based  centrality approaches. The node selectivity value is used for extraction nad ranking of keyword candidates. Furthermore, we expand extracted nodes to words-tuples with the highest in/out selectivity values. 

Selectivity-based extraction does not require linguistic knowledge while it is purely derived from statistical and structural information encompassed in the source text which is reflected into the structure of the network. Obtained average $F1$ score for the set of extracted keyword candidates is 24,63\%, and average $F2$ score is 21,19\%. Expansion of obtained candidates to words-tuples increased average $F1$ score to 25,9\% and $F2$ score to 24,47\%, which is comparable to the results on the same dataset achieved by supervised and unsupervised methods.

Our results imply that the structure of the network can be applied on Croatian keyword extraction task with many possible improvements. This should be thoroughly examined in the future work, which will cover: a) evaluation - consider all flective word forms; consider different matching strategies - exact, fuzzy, partOf match; b) text types - consider texts of different length, genre and topics; c) multitopic - compare isolate document extraction vs. multitopic extraction (one network constructed for the whole collection); d) other languages - test on standard English (and other) datasets; e) compare with other centrality motivated extraction approaches; f) longer keywords candidate sets - construct keyword sequences up to length 3; g) entity extraction - test weather entities can be extracted from complex networks.

\bibliographystyle{splncs}

\end{document}